%% file: egpaper_final.tex
\documentclass[10pt,twocolumn,letterpaper]{article}

\usepackage{iccv}
\usepackage{times}
\usepackage{epsfig}
\usepackage{graphicx}
\usepackage{amsmath}
\usepackage{amssymb}
\usepackage{multirow}
\usepackage{adjustbox}
\usepackage{booktabs}
\usepackage{siunitx}
\usepackage{textcomp}
\usepackage{float}

\usepackage[breaklinks=true,bookmarks=false]{hyperref}

\iccvfinalcopy 

\def\iccvPaperID{1349} 

\ificcvfinal\fi

\begin{document}

\title{Geometry-based Distance Decomposition for Monocular 3D Object Detection}

\author{
\parbox{16cm}{
\centering
{\large Xuepeng Shi$^{1}$ \ Qi Ye$^{3}$ \ Xiaozhi Chen$^{2}$ \ Chuangrong Chen$^{2}$ \ Zhixiang Chen$^{1}$ \ Tae-Kyun Kim$^{1,4}$}\\
{\normalsize
$^1$ Imperial College London \quad $^2$ DJI \quad $^3$ Zhejiang University\\
$^4$ Korea Advanced Institute of Science and Technology\\
}
{\tt\small \{x.shi19, zhixiang.chen, tk.kim\}@imperial.ac.uk \quad \{qi.ye\}@zju.edu.cn}
}
}

\maketitle
\ificcvfinal\thispagestyle{empty}\fi

\begin{abstract}
Monocular 3D object detection is of great significance for autonomous driving but remains challenging. The core challenge is to predict the distance of objects in the absence of explicit depth information. Unlike regressing the distance as a single variable in most existing methods, we propose a novel geometry-based distance decomposition to recover the distance by its factors. The decomposition factors the distance of objects into the most representative and stable variables, i.e. the physical height and the projected visual height in the image plane. Moreover, the decomposition maintains the self-consistency between the two heights, leading to robust distance prediction when both predicted heights are inaccurate. The decomposition also enables us to trace the causes of the distance uncertainty for different scenarios. Such decomposition makes the distance prediction interpretable, accurate, and robust. Our method directly predicts 3D bounding boxes from RGB images with a compact architecture, making the training and inference simple and efficient. The experimental results show that our method achieves the state-of-the-art performance on the monocular 3D Object Detection and Bird’s Eye View tasks of the KITTI dataset, and can generalize to images with different camera intrinsics~\footnote{https://github.com/Rock-100/MonoDet}.
\end{abstract}

\input{./Content/Introduction}
\input{./Content/Related_Work}
\input{./Content/Proposed_MonoRCNN}
\input{./Content/Experiments}

\section{Conclusion}
We have proposed a novel geometry-based distance decomposition which makes the distance prediction interpretable, accurate, and robust. Our method directly predicts 3D bounding boxes from RGB images with a compact architecture, thus is simple and efficient. The experimental results show that our method achieves the SOTA performance on the monocular 3D Object Detection and Bird’s Eye View tasks of the KITTI dataset, and can generalize to images with different camera intrinsics.

{\small
\bibliographystyle{ieee_fullname}
\bibliography{egbib}
}

\input{./Content/Materials}

\end{document}

%% file: Content/Introduction.tex
\section{Introduction}

\begin{figure}
\centering
\includegraphics[width=0.48\textwidth]{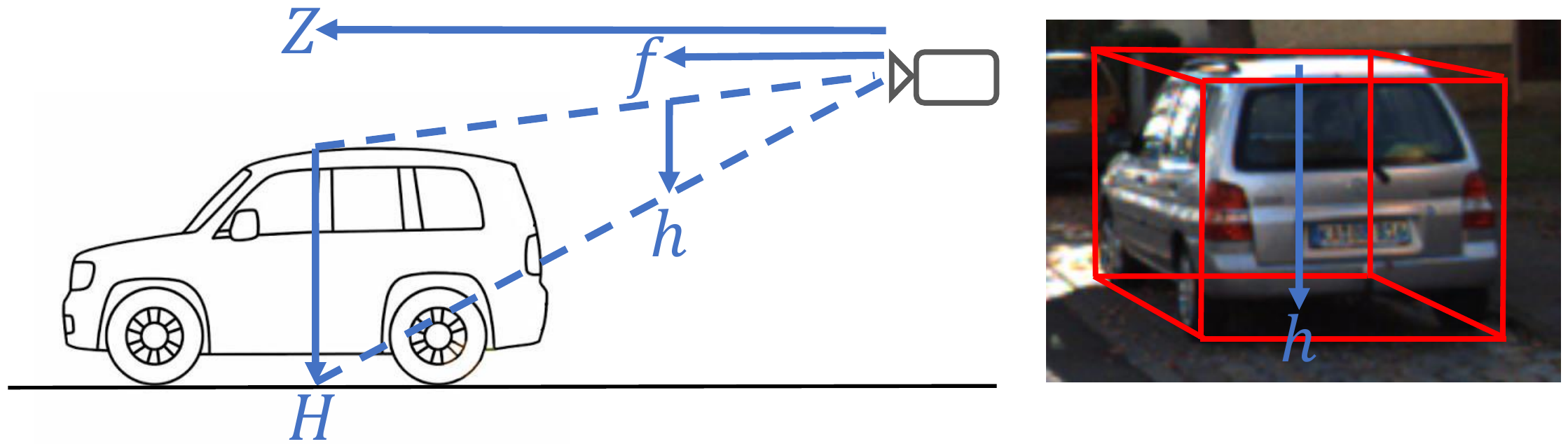}
\caption{Our distance decomposition is based on the imaging geometry~\cite{Hartley2000} of a pinhole camera. The distance from the center of an object to the camera, denoted as $Z$, can be calculated by $Z = \frac{f H}{h}$, where $f$ denotes the focal length of the camera, $H$ denotes the physical height of the object, and $h$ denotes the length of the projected central line (PCL). The PCL represents the projection of the vertical line at the center of the 3D bounding box. This equation shows the distance of objects is determined by the physical height and projected visual height in the image plane.}
\label{fig:Decomposition}
\end{figure}

Object detection is a fundamental and challenging problem in computer vision. With the emergence of deep learning~\cite{DBLP:journals/corr/SimonyanZ14a,DBLP:conf/cvpr/HeZRS16}, 2D object detection has achieved great progress in the past years~\cite{DBLP:conf/cvpr/GirshickDDM14,DBLP:conf/iccv/Girshick15,DBLP:conf/nips/RenHGS15,DBLP:conf/eccv/LiuAESRFB16,DBLP:conf/iccv/LinGGHD17,DBLP:conf/cvpr/LinDGHHB17}. However, it is yet insufficient for applications requiring 3D spatial information like autonomous driving. 3D object detection, which detects objects as 3D bounding boxes, has drawn much attention. Comparing with 3D object detection methods relying on expensive LiDAR sensors to provide depth information~\cite{DBLP:conf/cvpr/YangLU18,DBLP:conf/cvpr/ShiWL19,DBLP:conf/cvpr/ZhouT18,DBLP:conf/cvpr/LangVCZYB19}, monocular 3D object detection~\cite{DBLP:conf/cvpr/ChenKZMFU16,DBLP:conf/bmvc/RoddickKC19} infers depth from monocular images with low computation and energy cost. The core challenge of monocular 3D object detection is inferring the distance of objects in the absence of explicit depth information. Given the visual appearance of an object, its spatial location can be inferred based on the imaging geometry~\cite{Hartley2000} as an inverse problem. Thus, the priors of object physical size, scene layout, and the imaging process of cameras are essential to exploit to recover the distance.

On one hand, such geometric priors have been exploited to predict the pose or distance of objects by their factors. In 6D object pose estimation, PVNet~\cite{DBLP:conf/cvpr/PengLHZB19} and SegDriven~\cite{DBLP:conf/cvpr/HuHFS19} regress the 2D keypoints of objects. In category-level
6D object pose and size estimation, NOCS~\cite{DBLP:conf/cvpr/Wang0HVSG19} uses the normalized object coordinate space map. In stereo 3D object detection, Stereo R-CNN~\cite{DBLP:conf/cvpr/0001CS19} uses sparse 2D keypoints, yaw angle, object physical size, and a region-based photometric alignment using the left and right RoIs. These works~\cite{DBLP:conf/cvpr/PengLHZB19,DBLP:conf/cvpr/HuHFS19,DBLP:conf/cvpr/0001CS19,DBLP:conf/cvpr/Wang0HVSG19} recover the pose or distance by several factors, such as 2D keypoints, 2D bounding boxes, and object physical size, which achieves interpretable and robust pose or distance estimation. 

On the other hand, most monocular 3D object detection methods deal with the challenging distance prediction by regressing it as a single variable. The learning-based methods~\cite{DBLP:journals/corr/abs-1904-07850,DBLP:conf/cvpr/ChenTSL20,DBLP:conf/iccv/SimonelliBPLK19,DBLP:conf/eccv/SimonelliBPRK20} directly learn a mapping from input images to the distance. The pseudo-LiDAR based methods~\cite{DBLP:conf/cvpr/WangCGHCW19,DBLP:conf/iccv/MaWLZOF19,DBLP:conf/eccv/YeDSLTFDW20} first regress the depth map of an input image and then predict the distance of objects with the depth map. The 3D-anchor-based methods~\cite{DBLP:conf/iccv/Brazil019,DBLP:conf/eccv/BrazilPLS20,DBLP:conf/cvpr/DingHYWSLL20a} break the distance prediction into the region proposal and the offset regression. The only exceptions are~\cite{DBLP:conf/cvpr/MousavianAFK17,DBLP:journals/corr/abs-1906-08070,DBLP:conf/eccv/LiZLC20}, which recover the distance by minimizing the re-projection error between 3D bounding boxes and 2D bounding boxes or 2D keypoints. 
However, \cite{DBLP:conf/cvpr/MousavianAFK17,DBLP:journals/corr/abs-1906-08070,DBLP:conf/eccv/LiZLC20} still lag behind those methods which regress the distance as a single variable.

Aiming to close this gap, we propose a novel geometry-based distance decomposition to recover the distance by its factors. Different from~\cite{DBLP:conf/cvpr/MousavianAFK17,DBLP:journals/corr/abs-1906-08070,DBLP:conf/eccv/LiZLC20}, we abstract objects as vertical lines at the center of 3D bounding boxes, and their visual projection as the projection of these vertical lines, then recover the distance by them based on the imaging geometry~\cite{Hartley2000}, as shown in Fig.~\ref{fig:Decomposition}. The decomposition is designed to be \textit{as simple as possible, yet effective and efficient} to extract the most representative and stable factors of the distance of objects, i.e., the physical height and the projected visual height. The advantages of the decomposition are four-fold. 1) It makes the distance prediction interpretable. The physical height can be interpreted as an intrinsic attribute of objects, and the visual height can be interpreted as the extrinsic position in a scene. 2) The physical height and projected visual height are easy to estimate as revealed by our observations. 3) The decomposition maintains the self-consistency between the two heights, leading to robust distance prediction when both predicted heights are inaccurate. 4) The decomposition enables us to trace and interpret the causes of the distance uncertainty for different scenarios, by introducing an uncertainty-aware regression loss for the decomposed variables. In addition, our method can generalize to images with different camera intrinsics, because it reasons the distance only by the local information of objects and decouples the focal length from the distance prediction. This generalization ability is crucial to facilitate the deployment of the machine learning models of monocular 3D vision~\cite{DBLP:conf/cvpr/FacilUZMBC19}.

The contributions of our method are summarized below:
\begin{enumerate}
\item A novel geometry-based distance decomposition makes the distance prediction interpretable, accurate and robust.
\item Based on the decomposition, our method originally traces the causes of the distance uncertainty.
\item Our method directly predicts 3D bounding boxes from RGB images with a compact architecture, making the training and inference simple and efficient.
\item Our method achieves the state-of-the-art (SOTA) performance on the monocular 3D Object Detection and Bird’s Eye View tasks of the KITTI dataset~\cite{DBLP:conf/cvpr/GeigerLU12}, and can adapt to images with different camera intrinsics.
\end{enumerate}

%% file: Content/Related_Work.tex
\begin{figure*}
\centering
\includegraphics[width=0.9\textwidth]{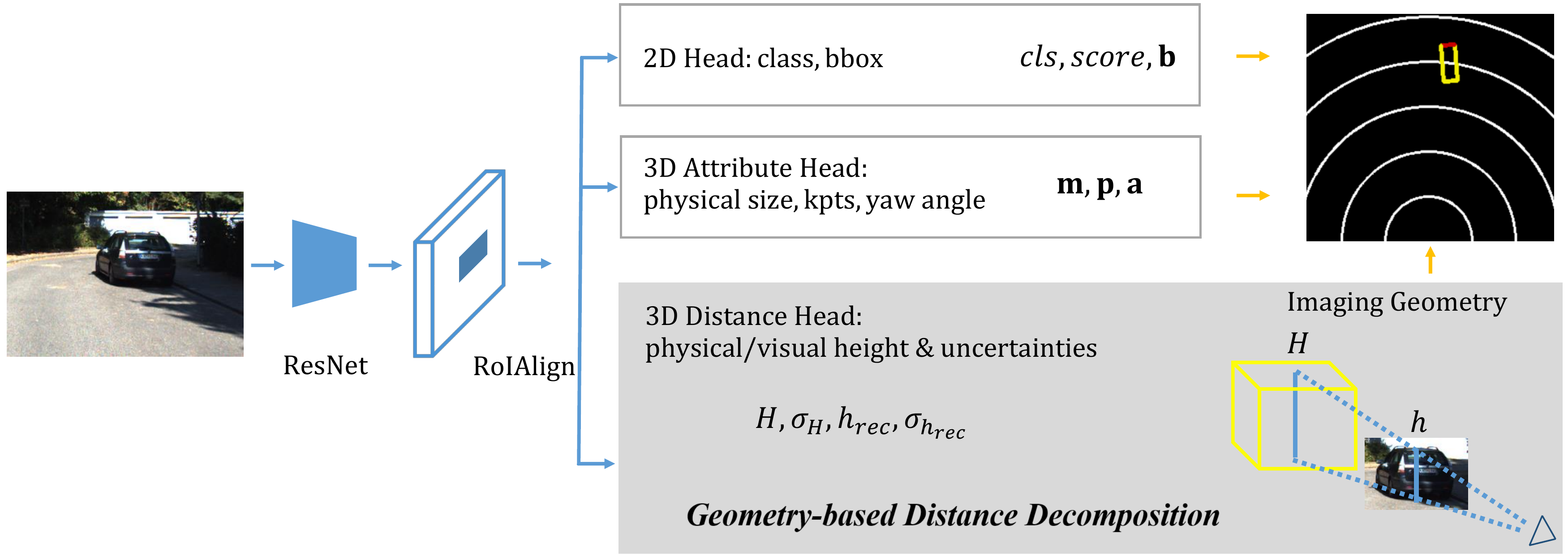}
\caption{The main architecture of MonoRCNN. MonoRCNN is built upon Faster R-CNN~\cite{DBLP:conf/nips/RenHGS15} and adds the carefully designed 3D distance head. The 3D distance head is based on our geometry-based distance decomposition. Specifically, our method regresses $H$, $h_{rec} = \frac{1}{h}$, and their uncertainties, then recovers the distance by $Z = f H h_{rec}$.  \textcolor[RGB]{100,149,237}{Blue} arrows represent operations in the network during training and inference, and \textcolor[RGB]{255,165,0}{orange} arrows represent operations to recover 3D bounding boxes during inference.
}
\label{fig:Framework}
\end{figure*}

\section{Related Work}

\subsection{2D Object Detection}
2D object detection has achieved sustainable improvements~\cite{DBLP:conf/cvpr/GirshickDDM14,DBLP:conf/iccv/Girshick15,DBLP:conf/nips/RenHGS15,DBLP:conf/eccv/LiuAESRFB16,DBLP:conf/iccv/LinGGHD17,DBLP:conf/cvpr/LinDGHHB17} in the past years. Notably, the two-stage frameworks, such as Faster R-CNN~\cite{DBLP:conf/nips/RenHGS15} and Mask R-CNN~\cite{DBLP:conf/iccv/HeGDG17}, achieve dominated performance on several challenging datasets~\cite{DBLP:conf/eccv/LinMBHPRDZ14,DBLP:conf/cvpr/GeigerLU12}. Feature pyramid networks (FPN)~\cite{DBLP:conf/cvpr/LinDGHHB17} has also been proposed to improve the 2D object detection performance. We adopt Faster R-CNN~\cite{DBLP:conf/nips/RenHGS15} with FPN~\cite{DBLP:conf/cvpr/LinDGHHB17} as our 2D object detection framework, because of its high accuracy and flexibility. 

\subsection{Monocular 3D Object Detection}

Most monocular 3D object detection methods deal with the challenging distance prediction by regressing it as a single variable. The learning-based methods~\cite{DBLP:journals/corr/abs-1904-07850,DBLP:conf/cvpr/ChenTSL20,DBLP:conf/iccv/SimonelliBPLK19,DBLP:conf/eccv/SimonelliBPRK20} directly regress the distance of objects by adding distance branches to 2D object detectors, which are simple and efficient. The pseudo-LiDAR-based methods~\cite{DBLP:conf/cvpr/WangCGHCW19,DBLP:conf/iccv/MaWLZOF19,DBLP:conf/eccv/YeDSLTFDW20} first predict the depth map of an input image using an external monocular depth estimator, then predict the distance of objects from the estimated depth map using a point-cloud-based 3D object detector. Though the explicit depth cues from the the estimated depth map can ease the distance prediction, the generalization of these methods is bounded by that of the monocular depth estimators~\cite{DBLP:journals/corr/abs-2012-05796}. The 3D-anchor-based methods~\cite{DBLP:conf/iccv/Brazil019,DBLP:conf/eccv/BrazilPLS20,DBLP:conf/cvpr/DingHYWSLL20a} extend the 2D anchor boxes~\cite{DBLP:conf/nips/RenHGS15} to the 3D anchor boxes by supplementing 3D bounding box templates, then predict the transformations from the 3D anchor boxes to the ground-truth 3D bounding boxes. The 3D anchor boxes can ease the distance learning. Unlike regressing the distance as a single variable in these methods, we propose a novel geometry-based distance decomposition to recover the distance by its factors.

A few works~\cite{DBLP:conf/cvpr/MousavianAFK17,DBLP:journals/corr/abs-1906-08070,DBLP:conf/eccv/LiZLC20} predict the distance of objects by its factors. 
Deep3Dbox~\cite{DBLP:conf/cvpr/MousavianAFK17} abstracts objects as 3D bounding boxes and their visual projection as the four boundaries of projected 3D bounding boxes, then recovers the distance by minimizing the re-projection error between the four boundaries of projected 3D bounding boxes and 2D bounding boxes. The keypoint-based methods~\cite{DBLP:journals/corr/abs-1906-08070,DBLP:conf/eccv/LiZLC20} abstract objects as 3D bounding boxes and their visual projection as the eight projected corners of 3D bounding boxes, then recovers the distance by minimizing the re-projection error between the eight projected corners of 3D bounding boxes and the predicted eight projected corners. However, \cite{DBLP:conf/cvpr/MousavianAFK17,DBLP:journals/corr/abs-1906-08070,DBLP:conf/eccv/LiZLC20} still lag behind those methods which regress the distance as a single variable. Similar to~\cite{DBLP:conf/cvpr/MousavianAFK17,DBLP:journals/corr/abs-1906-08070,DBLP:conf/eccv/LiZLC20}, our method also recovers the distance by its factors, but our decomposition is more simple and effective.

\subsection{Geometry-based Object Pose Estimation}
Geometric priors have been exploited to predict the pose or distance of objects by their factors. In 6D object pose estimation, PVNet~\cite{DBLP:conf/cvpr/PengLHZB19} and SegDriven~\cite{DBLP:conf/cvpr/HuHFS19} regress 2D keypoints of objects, then optimize the estimation of the 6D pose by solving a Perspective-n-Point (PnP) problem. In category-level
6D object pose and size estimation, NOCS~\cite{DBLP:conf/cvpr/Wang0HVSG19} uses the normalized object coordinate space map together with the depth
map in a pose fitting algorithm, to estimate the 6D pose and physical size of unseen objects. In stereo 3D object detection, stereo R-CNN~\cite{DBLP:conf/cvpr/0001CS19} first calculates the coarse distance from sparse 2D keypoints, yaw angle, and object physical size, then recovers the accurate distance by a region-based photometric alignment using the left and right RoIs. Inspired by these works, we propose a geometry-based distance decomposition for monocular 3D object detection, to recover the distance by its factors.

\subsection{Uncertainty Estimation}
There are two seminal works~\cite{DBLP:conf/nips/KendallG17,DBLP:conf/cvpr/KendallGC18} exploring uncertainties in deep learning for computer vision. The uncertainty-aware regression loss~\cite{DBLP:conf/cvpr/KendallGC18} enables networks to re-balance samples and re-focus on more reasonable samples, which improves the overall accuracy. MonoLoco~\cite{DBLP:conf/iccv/BertoniKA19}, MonoPair~\cite{DBLP:conf/cvpr/ChenTSL20} and UR3D~\cite{DBLP:conf/eccv/ShiCK20} regress the distance of objects with the uncertainty-aware regression loss~\cite{DBLP:conf/cvpr/KendallGC18} to improve the distance regression accuracy. MonoDIS~\cite{DBLP:conf/iccv/SimonelliBPLK19} proposes a self-supervised confidence score to re-sort the predicted 3D bounding boxes. Kinematic3D~\cite{DBLP:conf/eccv/BrazilPLS20} proposes a self-balancing 3D confidence loss to both improve the 3D box regression accuracy and re-sort the predicted 3D bounding boxes. \cite{DBLP:conf/iccv/BertoniKA19,DBLP:conf/cvpr/ChenTSL20,DBLP:conf/eccv/ShiCK20,DBLP:conf/iccv/SimonelliBPLK19,DBLP:conf/eccv/BrazilPLS20} directly apply the uncertainty-aware losses to the distance. Instead, we apply the uncertainty-aware regression loss~\cite{DBLP:conf/cvpr/KendallGC18} to the decomposed variables of the distance, which enables us to trace the causes of distance uncertainty for different scenarios. 

%% file: Content/Proposed_MonoRCNN.tex
\section{Proposed MonoRCNN}
We first present the basic framework, then two 3D-related detection heads, i.e., the 3D distance head and 3D attribute head. We detail the geometry-based distance decomposition and uncertainty-aware regression in the 3D distance head. We term our method as MonoRCNN, and the main architecture is illustrated in Fig.~\ref{fig:Framework}.

\subsection{Basic Framework}
We address monocular 3D object detection, which predicts the 3D bounding boxes of objects from monocular RGB images. Two common assumptions~\cite{DBLP:conf/cvpr/GeigerLU12} are 1) only considering the yaw angle of 3D bounding boxes and setting the roll and pitch angle as zero, 2) per-image camera intrinsics are available both during training and inference. For a given RGB image, MonoRCNN reports all objects within concerned categories, and the output for each object is
\begin{enumerate}
\item class label $\mathit{cls}$ and confidence $\mathit{score}$,
\item 2D bounding box represented by the top-left and bottom-right corners, denoted as $\mathbf b = (x_1, y_1, x_2, y_2)$,
\item the 2D projected center of the 3D bounding box, denoted as $\mathbf p = (p_1, p_2)$,
\item the physical size of the 3D bounding box, denoted as $\mathbf m = (W, H, L)$, where $W, H, L$ are the physical width, height, and length, respectively,
\item the yaw angle of the 3D bounding box, denoted as $\mathbf a = (\sin(\theta), \cos(\theta))$, where $\theta$ is the allocentric pose of the 3D bounding box,
\item the distance of the center of the 3D bounding box, denoted as $Z$.
\end{enumerate}

MonoRCNN predicts the 3D center $(p_1, p_2, Z)$ in pixel coordinates, and convert it to camera coordinates using the projection matrix $\mathbf{P}$ during inference, formulated as
\begin{equation}
\begin{aligned}
\begin{bmatrix}
    p_1 \cdot Z \\
    p_2 \cdot Z \\
    Z  \\
\end{bmatrix}_\text{P} =
\mathbf{P} \cdot 
\begin{bmatrix}
    x  \\
    y  \\
    z \\
    1 \\
\end{bmatrix}_\text{C}.
\end{aligned}
\end{equation}
For the yaw angle prediction, MonoRCNN predicts $\sin(\theta)$ and $\cos(\theta)$, and convert them to $\theta$ during inference.

MonoRCNN is built upon Faster R-CNN~\cite{DBLP:conf/nips/RenHGS15}. We use a ResNet-50~\cite{DBLP:conf/cvpr/HeZRS16} with FPN~\cite{DBLP:conf/cvpr/LinDGHHB17} as the backbone and RoIAlign~\cite{DBLP:conf/iccv/HeGDG17} to extract the crops of object features. For the training and inference of the 2D object detection network, we follow the pipelines in ~\cite{DBLP:conf/nips/RenHGS15,DBLP:conf/iccv/HeGDG17}. To adapt to monocular 3D object detection, we add the 3D distance head and 3D attribute head.

\begin{figure}
\centering
\includegraphics[width=0.475\textwidth]{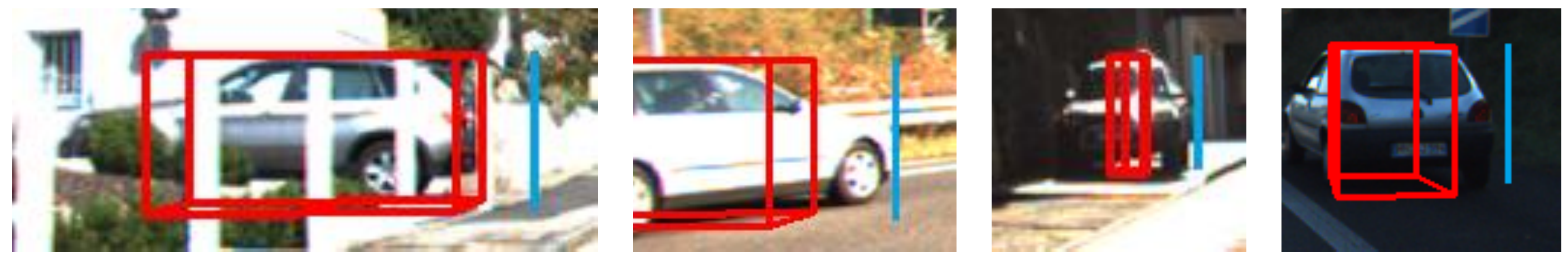}
\caption{Comparisons between the predicted eight projected corners (\textcolor{red}{red boxes}) and predicted visual height (\textcolor[RGB]{100,149,237}{blue lines}). Predicting the eight projected corners fails under challenging cases, such as occlusion, truncation, and extreme lighting conditions, while predicting the visual height is more simple and robust. The images are from the val subset of the KITTI validation split~\cite{DBLP:conf/nips/ChenKZBMFU15}.}
\label{fig:Kpts}
\vspace{-0.4cm}
\end{figure}

\subsection{3D Distance Head}

\subsubsection{Geometry-based Distance Decomposition}
The 3D distance head recovers the distance of objects and is based on our geometry-based distance decomposition. Specifically, we decompose the distance of an object $Z$, into the physical height $H$, and the reciprocal of the projected visual height $h_{rec} = \frac{1}{h}$, which is formulated as 
\begin{equation}
\begin{aligned}
Z = \frac{f H}{h} = f H h_{rec},
\end{aligned}
\end{equation}
where $f$ denotes the focal length of the camera. We regress $H$ and $h_{rec}$ separately and recover $Z$ by them. 

The decomposition makes the distance prediction interpretable. $H$ can be interpreted as an intrinsic attribute of objects, the estimation of which can be regarded as a fine-grained object classification problem. While given an object, $h_{rec}$ can be interpreted as the extrinsic position in a scene, the estimation of which is a 2D regression problem in the image plane.

The physical height and visual height required by our decomposition are easy to estimate. Predicting the eight projected corners of 3D bounding boxes~\cite{DBLP:journals/corr/abs-1906-08070,DBLP:conf/eccv/LiZLC20} is challenging due to the occlusion, truncation, yaw angle variations, and extreme lighting conditions. As shown in Fig.~\ref{fig:Kpts}, the visual height prediction is accurate in different challenging cases but the projected corner prediction fails.
Moreover, the physical height is the simplest and the most stable variable among the physical size, according to the prediction error of the physical size of cars on the val subset of the KITTI validation split~\cite{DBLP:conf/nips/ChenKZBMFU15}, shown in Tab.~\ref{tab:Size_error}. The mean prediction error of the physical height is much smaller than that of the physical length. In addition, the prediction error of the physical length and width are influenced by the yaw angle due to single-view ambiguity, while the prediction error of the physical height is not. Our decomposition only uses the physical height, instead of the full physical size~\cite{DBLP:journals/corr/abs-1906-08070,DBLP:conf/eccv/LiZLC20}, to recover the distance, which improves the distance prediction accuracy.

\begin{table}
\begin{center}
\begin{adjustbox}{max width=\textwidth}
\begin{tabular}{l | c c c}
\toprule[1pt]
\multirow{2}{*}{ } & 
\multicolumn{3}{c}{Mean prediction error (meters)~\textdownarrow}  \\ 
& S \& F \& B & S & F \& B \\ 
\midrule[0.5pt]
Length & 0.293 & 0.276 & 0.296 \\
Width & 0.071 & 0.078 & 0.070 \\
Height & 0.078 & 0.076 & 0.078 \\
\bottomrule[1pt]
\end{tabular}
\end{adjustbox}
\end{center}
\caption{The prediction error of the physical size within different yaw angle ranges on the val subset of the KITTI validation split~\cite{DBLP:conf/nips/ChenKZBMFU15}. `S', `F', and `B' denote cars whose visible parts are side, front, and back, respectively.}
\label{tab:Size_error}
\vspace{-0.4cm}
\end{table}

\begin{figure}
\centering
\includegraphics[width=0.45\textwidth]{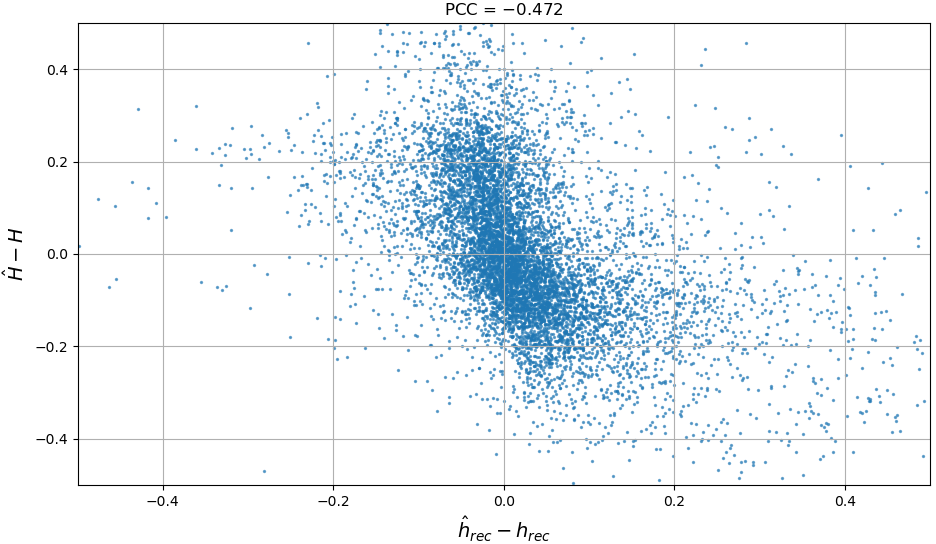}
\caption{The learned correlation between $H$ and $h_{rec}$ on the val subset of the KITTI validation split~\cite{DBLP:conf/nips/ChenKZBMFU15}. The Pearson correlation coefficient (PCC) of $\hat{h}_{rec} - h_{rec}$ and $\hat{H} - H$ is \num{-0.472}. Errors are normalized to $[\num{-0.5}, \num{0.5}]$.
}
\label{fig:Corr}
\vspace{-0.4cm}
\end{figure}

The decomposition can also maintain the self-consistency during inference, leading to robust distance prediction when both predicted heights are inaccurate. 
With the objective of predicting the distance, the neural network can learn the correlation between $H$ and $h_{rec}$ during training, then the learned correlation can serve as the self-consistency during inference. This correlation is detailed as below. In the context of monocular 3D object detection task, the physical height $H$ is a constant for a given object. The distance of the object to the camera $Z$ can be modeled as a random variable since the object can appear in different locations in a scene. Similarly, the reciprocal of the length of the PCL of the object $h_{rec}$ is also a random variable. While the variable $Z$ is random for a specified object, we notice that this variable is expected to follow a same distribution for different objects, denoted as $\mathcal{D}$. This is because the distribution of the location of an object is irrelevant to its fine-grained object type. For example, the spatial positions of cars on a street are not influenced by their car types. We formulate this as  
\begin{equation}
\begin{aligned}
Z = f H h_{rec} \sim \mathcal{D}.
\label{eq:distribution}
\end{aligned}
\end{equation}
By taking expectation on Eq.~\eqref{eq:distribution}, we have
\begin{equation}
\begin{aligned}
H \mathbb{E}[h_{rec}] = \frac{\mathbb{E}[Z]}{f}. 
\label{eq:product}
\end{aligned}
\end{equation}
Eq.~\eqref{eq:product} shows that, for the training labels of different objects, the product between their $H$ and their expectation of $h_{rec}$ is a constant. In other words, for different objects, the expectation of $h_{rec}$ decreases with the increase of $H$. \textit{Intuitively speaking, the larger the physical height of an object, the larger the average projected visual height of this object}. This is the correlation between the training labels of $H$ and $h_{rec}$. The neural network can learn this correlation during training, as shown in Fig.~\ref{fig:Corr}. During inference, if the predicted $H$ is larger than the groundtruth, the learned correlation pushes the predicted $h_{rec}$ to become smaller on average, and vice versa. Thus, our method can recover the accurate distance $Z$ with the inaccurate $H$ and $h_{rec}$, i.e., maintain the self-consistency during inference.

\subsubsection{Uncertainty-aware Regression}

Based on the decomposition, we further trace and interpret the causes of the distance uncertainty for different scenarios. We modify the uncertainty-aware regression loss~\cite{DBLP:conf/cvpr/KendallGC18} to regress $H$ and $h_{rec}$. The loss functions for $H$ and $h_{rec}$ can be formulated as

\begin{equation}
\begin{aligned}
L_{H} = \frac{L_{1}(\hat{H}, H)}{\sigma_H} + \lambda_{H} log(\sigma_H),
\end{aligned}
\label{eq:loss_dist_H}
\end{equation}
\begin{equation}
\begin{aligned}
L_{h_{rec}} = \frac{L_{1}(\hat{h}_{rec}, h_{rec})}{\sigma_{h_{rec}}} + \lambda_{h_{rec}} log(\sigma_{h_{rec}}),
\end{aligned}
\label{eq:loss_dist_h}
\end{equation}
where $\hat{H}$ and $\hat{h}_{rec}$ are the groundtruths, $H$ and $h_{rec}$ are the predictions, $\lambda_{H}$ and $\lambda_{h_{rec}}$ are the positive parameters to balance the uncertainty terms, and $\sigma_H$ and $\sigma_{h_{rec}}$ are the learnable variables of uncertainties. 

\begin{figure}
\centering
\includegraphics[width=0.48\textwidth]{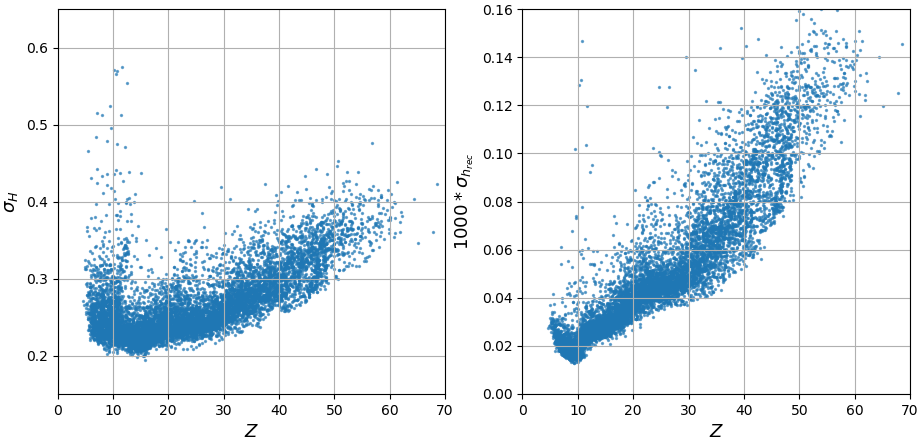}
\caption{The uncertainties of $H$ (left) and $h_{rec}$ (right) vs. distance $Z$ (meters) on the val subset of the KITTI validation split~\cite{DBLP:conf/nips/ChenKZBMFU15}. 
}
\label{fig:Uncertainty}
\vspace{-0.4cm}
\end{figure}

In Fig.~\ref{fig:Uncertainty}, we show the uncertainties of the physical height and the projected visual height, i.e., $\sigma_H$ and $\sigma_{h_{rec}}$, for objects at different distances on the val subset of the KITTI validation split~\cite{DBLP:conf/nips/ChenKZBMFU15}. 
For both $H$ and $h_{rec}$, the uncertainties first decrease and then increase when objects go far away from the camera. At a close distance, the high uncertainties are mainly caused by the truncated views of objects. It is hard to make accurate prediction with partial observations. At a far distance, the high uncertainties are mainly caused by the coarse views of objects with fewer pixels representing them in images. The truncated and coarse views result in comparable increases for the uncertainties of $H$. However, $h_{rec}$ is with noticeable higher uncertainties for the coarse views than the truncated views. In other words, the accuracy of the distance estimation for faraway objects is heavily influenced by the accuracy of $h_{rec}$.

$\sigma_{h_{rec}}$ is more discriminative for objects at different distance than $\sigma_{H}$, and $f H \sigma_{h_{rec}}$ can represent the distance uncertainty. We use $\frac{score}{f H \sigma_{h_{rec}}}$, instead of $score$, to sort the predicted boxes to improve the 3D object detection accuracy.

\subsection{3D Attribute Head}
3D attribute head predicts the physical size, yaw angle, and 2D keypoints, i.e., the projected center and corners of 3D bounding boxes. We use the $L_{1}$ loss to directly regress the physical size and yaw angle, formulated as
\begin{equation}
\begin{aligned}
L_{size} = L_1(\hat{\mathbf{m}}, \mathbf m),
\end{aligned}
\end{equation}
\begin{equation}
\begin{aligned}
L_{yaw} = L_1(\hat{\mathbf{a}}, \mathbf a),
\end{aligned}
\end{equation}
where $\hat{\mathbf{m}}$ and $\hat{\mathbf{a}}$ are the groundtruths, and $\mathbf{m}$ and $\mathbf{a}$ are the predictions. For the keypoint regression, we normalize the keypoints by their proposal size. Let $(x_1, y_1, x_2, y_2)$ denote the top-left and bottom-right corners of a proposal, and $\hat{\mathbf{p}} = (\hat{p}_1, \hat{p}_2)$ and $\mathbf{p} = (p_1, p_2)$ denote the groundtruth keypoint and the predicted keypoint, respectively. Let $\hat{\mathbf{t}}$ and $\mathbf{t}$ denote the normalized groundtruth keypoint and the normalized predicted keypoint, respectively, and $\hat{\mathbf{t}}$ is defined as
\begin{equation}
\begin{aligned}
\hat{\mathbf{t}} = (\frac{\hat{p}_1 - x_1}{x_2 - x_1}, \frac{\hat{p}_2 - y_1}{y_2 - y_1}).
\end{aligned}
\end{equation}
The keypoint loss function can be formulated as
\begin{equation}
\begin{aligned}
L_{kpt} = L_1(\hat{\mathbf{t}}, \mathbf t).
\end{aligned}
\end{equation}
During inference, we transform the normalized predicted keypoint $\mathbf{t}$ to the predicted keypoint $\mathbf{p}$. We only use the projected center of a 3D bounding box during inference, the losses of the eight projected corners are auxiliary losses during training.

\subsection{Overall Loss} 
The overall training loss function for the detection heads is 
\begin{equation}
\begin{aligned}
L =~& \lambda_{cls} L_{cls} + \lambda_{bbox} L_{bbox} + \lambda_{size} L_{size} +\\ &\lambda_{yaw} L_{yaw} + \lambda_{kpt} L_{kpt} + L_{H} + L_{h_{rec}},
\end{aligned}
\end{equation}
where $\lambda_{cls}$ is \num{1}, $\lambda_{bbox}$ is \num{1}, $\lambda_{size}$ is \num{3}, $\lambda_{yaw}$ is \num{5}, $\lambda_{kpt}$ is \num{5}, $\lambda_{H}$ is \num{0.25}, and $\lambda_{h_{rec}}$ is \num{1}.

\subsection{Implementation Details} 

The backbone of MonoRCNN is ResNet-50~\cite{DBLP:conf/cvpr/HeZRS16} with FPN~\cite{DBLP:conf/cvpr/LinDGHHB17} and is pre-trained on ImageNet~\cite{DBLP:conf/cvpr/DengDSLL009}. We extract ROI features from P2, P3, P4 and P5 of the backbone, as defined in~\cite{DBLP:conf/cvpr/LinDGHHB17}. We use five scale anchors of \{32, 64, 128, 126, 512\} with three ratios \{0.5, 1, 2\}, and tile the anchors on P4. Images are scaled to a fixed height of \num{512} pixels for both training and inference. During training the batch size is \num{4}, and the total iteration number is \num{1e5} and \num{2e5} on the training subset of the KITTI validation split~\cite{DBLP:conf/nips/ChenKZBMFU15} and the KITTI official test split~\cite{DBLP:conf/cvpr/GeigerLU12}, respectively. We adopt the step strategy to adjust the learning rate. The initial learning rate is \num{0.01} and reduced by \num{10} times after $60\%$, $80\%$, $90\%$ iterations. During training random mirroring is used as augmentation, and during inference no augmentation is used. We implement our method with PyTorch~\cite{DBLP:conf/nips/PaszkeGMLBCKLGA19} and Detectron2~\cite{wu2019detectron2}. All the experiments run on a server with $2.2$ GHz CPU and GTX Titan X.

%% file: Content/Experiments.tex
\section{Experiments}

\begin{figure*}
\begin{minipage}{0.5\linewidth}
\includegraphics[width=1.0\textwidth]{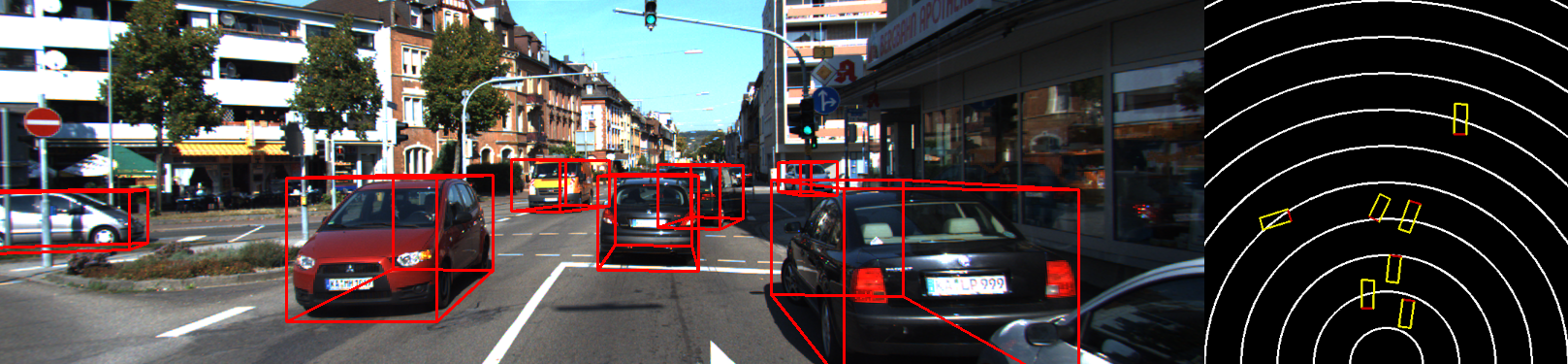}
\end{minipage}
\begin{minipage}{0.5\linewidth} 
\includegraphics[width=1.0\textwidth]{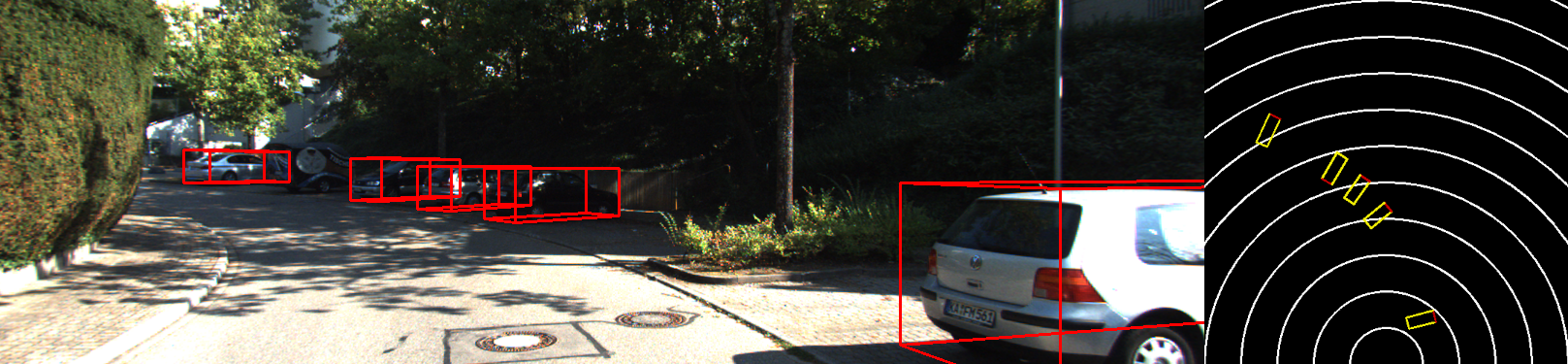}
\end{minipage}
\begin{minipage}{0.5\linewidth}
\includegraphics[width=1.0\textwidth]{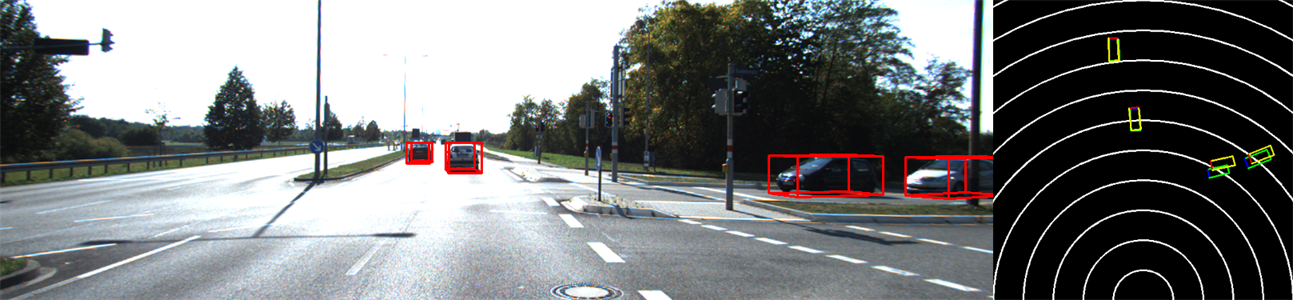}
\end{minipage}
\begin{minipage}{0.5\linewidth} 
\includegraphics[width=1.0\textwidth]{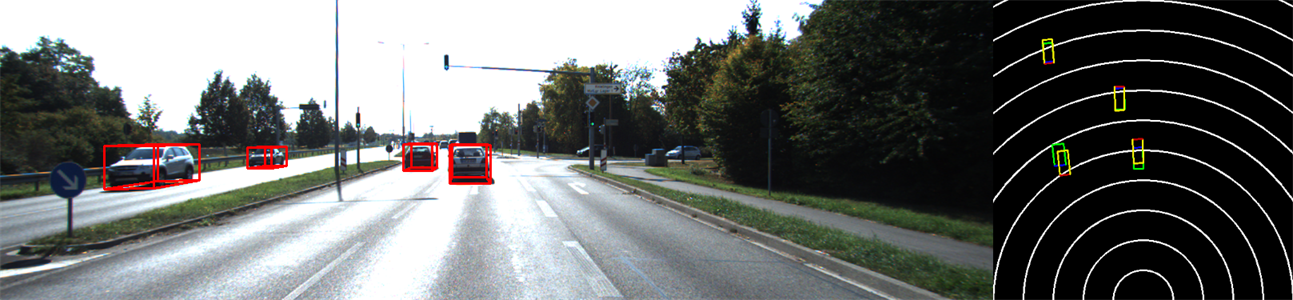}
\end{minipage}
\caption{\textbf{KITTI Examples}. We visualize qualitative examples of MonoRCNN on the KITTI test set~\cite{DBLP:conf/cvpr/GeigerLU12} (first row) and the val subset of the validation split~\cite{DBLP:conf/nips/ChenKZBMFU15} (second row). The \textcolor{red}{red} boxes in the image planes represent the 2D projections of the predicted 3D bounding boxes. The \textcolor{yellow}{yellow} / \textcolor{green}{green} boxes in the bird's eye view results represent the predictions and groundtruths, respectively, and the \textcolor{red}{red} / \textcolor{blue}{blue} lines indicate the yaw angle of cars. The radius difference between two adjacent white circles is \num{5} meters. All images are not used for training.
}
\label{fig:Example_K}
\end{figure*}

We first analyze the ablation studies and self-consistency on the KITTI validation split~\cite{DBLP:conf/nips/ChenKZBMFU15}. Then we comprehensively benchmark MonoRCNN on the KITTI official test dataset~\cite{DBLP:conf/cvpr/GeigerLU12}. We also present the cross-dataset test results using a nuScenes~\cite{DBLP:conf/cvpr/CaesarBLVLXKPBB20} cross-test set. Finally, we visualize qualitative examples on the KITTI dataset~\cite{DBLP:conf/cvpr/GeigerLU12} in Fig.~\ref{fig:Example_K}, and the nuScenes~\cite{DBLP:conf/cvpr/CaesarBLVLXKPBB20} cross-test set in Fig.~\ref{fig:Example_N}~\footnote{More qualitative examples can be found in the supplementary file.}.

\begin{table}
\begin{center}
\begin{adjustbox}{max width=\textwidth}
\begin{tabular}{l | c c }
\toprule[1pt]
\multirow{2}{*}{} & 
 \multicolumn{2}{c}{$\text{AP}|_{R_{40}}$~{[Easy~/~Mod~/~Hard]}~\textuparrow}  \\
& AP$_\text{3D}$  & AP$_\text{BEV}$ \\ 
\midrule[0.5pt]
L & $15.01$ / $10.52$ / $8.45$ & $21.03$ / $14.84$ / $11.44$  \\
\midrule[0.5pt]
K & $13.58$ / $8.96$ / $7.06$ & $19.39$ / $13.59$ / $10.54$  \\
\midrule[0.5pt]
D & $15.78$ / $10.97$ / $8.15$ & $22.06$ / $15.52$ / $11.82$  \\ 
D+U & $16.94$ / $12.00$ / $9.46$ & $24.60$ / $17.23$ / $13.38$  \\
D+U+S & $16.61$ / $13.19$ / $10.65$ & $25.29$ / $19.22$ / $15.30$   \\ 
\bottomrule[1pt]
\end{tabular}
\end{adjustbox}
\end{center}
\caption{\textbf{Ablation Studies} on the val subset of the KITTI validation split~\cite{DBLP:conf/nips/ChenKZBMFU15}. `L' means directly regressing the distance. `K' means using the eight projected corners and physical size to recover the distance, similar to~\cite{DBLP:journals/corr/abs-1906-08070,DBLP:conf/eccv/LiZLC20}. `D' means using our decomposition. `U' means adding the uncertainty-aware regression loss~\cite{DBLP:conf/cvpr/KendallGC18}. `S' means sorting the predicted boxes by $\frac{score}{f H \sigma_{h_{rec}}}$. 
}
\label{tab:Ablation}
\end{table}

\subsection{Datasets}
The KITTI dataset~\cite{DBLP:conf/cvpr/GeigerLU12} provides multiple widely used benchmarks for computer vision problems in autonomous driving. The Bird’s Eye View (BEV) and 3D Object Detection tasks are used to evaluate the 3D localization performance. These two tasks are characterized by \num{7481} training and \num{7518} test images with 2D and 3D annotations for cars, pedestrians, cyclists, etc. Each object is assigned with a difficulty level, i.e., easy, moderate or hard, based on its visual size, occlusion level and truncation degree. We conduct experiments on two common data splits, the val split~\cite{DBLP:conf/nips/ChenKZBMFU15} and the official test split~\cite{DBLP:conf/cvpr/GeigerLU12}. We only use the images from the left cameras for training. We report the $\text{AP}|_{R_{40}}$~\cite{DBLP:conf/iccv/SimonelliBPLK19} to compare the accuracy. We use the car class, the most representative class, and the official IoU criteria \num{0.7} for cars.

The nuScenes~\cite{DBLP:conf/cvpr/CaesarBLVLXKPBB20} 3D object detection task requires detecting \num{10} object classes in terms of full 3D bounding boxes, attributes and velocities. In this work, we focus on detecting the 3D bounding boxes of cars, to test the cross-dataset performance from KITTI~\cite{DBLP:conf/cvpr/GeigerLU12} to nuScenes~\cite{DBLP:conf/cvpr/CaesarBLVLXKPBB20}. We use the script~\footnote{https://github.com/nutonomy/nuscenes-devkit/blob/master/python-sdk/nuscenes/scripts/export\_kitti.py} for converting nuScenes data to KITTI format to generate a cross-test set. The cross-test set consists of \num{6019} front camera images from the official val subset. 

\begin{table*}
\begin{center}
\setlength\tabcolsep{12pt}
\begin{adjustbox}{max width=\textwidth}
\begin{tabular}{l | c | c | c c }
\toprule[1pt]
\multirow{2}{*}{Method} & 
\multirow{2}{*}{Input} &
\multirow{2}{*}{Time (ms)} & \multicolumn{2}{c}{$\text{AP}|_{R_{40}}$~{[Easy~/~Mod~/~Hard~]}~\textuparrow}  \\ 
& & & AP$_\text{3D}$  & AP$_\text{BEV}$ \\ 
\midrule[0.5pt]
ROI-10D (CVPR 19)~\cite{DBLP:conf/cvpr/ManhardtKG19}  & image $+$ depth & 200 & $4.32$ / $2.02$/ $1.46$ 	& $9.78$ / $4.91$/ $3.74$ \\
AM3D (ICCV 19)~\cite{DBLP:conf/iccv/MaWLZOF19} & image $+$ depth & 400 & $16.50$ / $10.74$ / $9.52$ & $25.03$ / $17.32$/ $\textcolor{blue}{14.91}$ \\ 
D$^4$LCN (CVPR 20)~\cite{DBLP:conf/cvpr/DingHYWSLL20a}  & image $+$ depth & 200 & $16.65$ / $11.72$ / $9.51$ & $22.51$ / $16.02$/ $12.55$ \\ 
DA-3Ddet (ECCV 20)~\cite{DBLP:conf/eccv/YeDSLTFDW20} & image $+$ depth & 400 & $16.77$ / $11.50$ / $8.93$ & $23.35$ / $15.90$/ $12.11$ \\
PatchNet (ECCV 20)~\cite{DBLP:conf/eccv/MaLXZZO20}  & image $+$ depth & 400 & $15.68$ / $11.12$ / $\textcolor{red}{10.17}$ & $22.97$ / $16.86$/ $\textcolor{red}{14.97}$ \\
Kinematic3D (ECCV 20)~\cite{DBLP:conf/eccv/BrazilPLS20} & image $+$ video & 120 & $\textcolor{red}{19.07}$ / $\textcolor{red}{12.72}$ / $9.17$ & $\textcolor{red}{26.69}$ / $\textcolor{blue}{17.52}$/ $13.10$ \\
\midrule[0.5pt]
FQNet (CVPR 19)~\cite{DBLP:conf/cvpr/LiuLXT019} & image & 500 & $2.77$ / $1.51$ / $1.01$ 	& $5.40$ / $3.23$/ $2.46$ \\ 
M3D-RPN (ICCV 19)~\cite{DBLP:conf/iccv/Brazil019} & image & 160 & $14.76$ / $9.71$ / $7.42$ 	& $21.02$ / $13.67$/ $10.23$ \\ 
MonoPair (CVPR 20)~\cite{DBLP:conf/cvpr/ChenTSL20} & image & 60 & $13.04$ / $9.99$ / $8.65$ 	& $19.28$ / $14.83$/ $12.89$ \\ 
MoVi-3D (ECCV 20)~\cite{DBLP:conf/eccv/SimonelliBPRK20}  & image & $-$ & $15.19$ / $10.90$ / $9.26$ 	& $22.76$ / $17.03$/ $14.85$ \\ 
RTM3D (ECCV 20)~\cite{DBLP:conf/eccv/LiZLC20}$^\dagger$  & image & 50 & $14.41$ / $10.34$ / $8.77$ 	& $19.17$ / $14.20$/ $11.99$ \\ 
MonoRCNN (Ours) & image & 70 & $\textcolor{blue}{18.36}$ / $\textcolor{blue}{12.65}$ / $\textcolor{blue}{10.03}$  & $\textcolor{blue}{25.48}$ / $\textcolor{red}{18.11}$/ $14.10$   \\ 
\bottomrule[1pt]
\end{tabular}
\end{adjustbox}
\end{center}
\caption{
\textbf{Comparisons on the KITTI benchmark~\cite{DBLP:conf/cvpr/GeigerLU12}}. `Input' means the input data modality used during training and inference. The inference time is reported from the official leaderboard with slight variances in hardware. \textcolor{red}{Red} / \textcolor{blue}{blue} indicate the best / second, respectively. $^\dagger$ denotes methods using additional images from right cameras for training.}
\label{tab:Test}
\end{table*}

\subsection{Ablation Studies}
We conduct ablation studies to examine how each proposed component affects the final performance. We evaluate the performance by first setting a baseline which utilizes our decomposition, then adding the uncertainty-aware loss~\cite{DBLP:conf/cvpr/KendallGC18}, finally sorting the predicted boxes by $\frac{score}{f H \sigma_{h_{rec}}}$, as shown in Tab.~\ref{tab:Ablation}. We also implement a model which directly regresses the distance to compare our method with the learning-based methods~\cite{DBLP:journals/corr/abs-1904-07850,DBLP:conf/cvpr/ChenTSL20,DBLP:conf/iccv/SimonelliBPLK19,DBLP:conf/eccv/SimonelliBPRK20}, and a model which recovers the distance by the eight projected corners and physical size to compare our method with the keypoint-based methods~\cite{DBLP:journals/corr/abs-1906-08070,DBLP:conf/eccv/LiZLC20}. From Tab.~\ref{tab:Ablation}, we can see:

1) The geometry-based distance decomposition is effective. Comparing `D' with `K', we can see the geometry-based distance decomposition outperforms the keypoint-based model by a large margin, which supports its effectiveness.

2) The uncertainty-aware regression loss~\cite{DBLP:conf/cvpr/KendallGC18} improves the accuracy. Comparing `D+U' with `D', we can see using the uncertainty-aware regression loss~\cite{DBLP:conf/cvpr/KendallGC18} for $H$ and $h_{rec}$ improves the accuracy.

3) Sorting by $\frac{score}{f H \sigma_{h_{rec}}}$ is effective. Comparing `D+U+S' with `D+U', we can see sorting by $\frac{score}{f H \sigma_{h_{rec}}}$ improves the accuracy, showing that $\frac{score}{f H \sigma_{h_{rec}}}$ represents the 3D prediction quality better than $score$.

\begin{table}
\begin{center}
\begin{adjustbox}{max width=\textwidth}
\begin{tabular}{l | c c }
\toprule[1pt]
\multirow{2}{*}{} & 
\multicolumn{2}{c}{$\text{AP}|_{R_{40}}$~{[Easy~/~Mod~/~Hard]}~\textuparrow}  \\
 & AP$_\text{3D}$  & AP$_\text{BEV}$ \\ 
\midrule[0.5pt]
K (P) & $13.58$ / $8.96$ / $7.06$ & $19.39$ / $13.59$ / $10.54$   \\
K (G) & $13.31$ / $9.51$ / $7.45$ & $21.13$ / $14.98$ / $11.44$   \\
\midrule[0.5pt]
Ours (P) & $16.94$ / $12.00$ / $9.46$ & $24.60$ / $17.23$ / $13.38$\\
Ours (G) & $14.45$ / $10.82$ / $8.62$ & $22.10$ / $16.45$ / $12.81$   \\
\bottomrule[1pt]
\end{tabular}
\end{adjustbox}
\end{center}
\caption{\textbf{Self-Consistency Comparisons} on the val subset of the KITTI validation split~\cite{DBLP:conf/nips/ChenKZBMFU15}. `K' means using the eight projected corners and physical size to recover the distance, similar to ~\cite{DBLP:journals/corr/abs-1906-08070,DBLP:conf/eccv/LiZLC20}. `P' means using the predicted physical height or size when recovering the distance, `G' means using the groundtruth instead.}
\label{tab:Consistency}
\end{table}

\subsection{Self-Consistency Comparisons}
We analyze the self-consistency as shown in Tab.~\ref{tab:Consistency}. From `Ours', we can see that, the accuracy decreases if the predicted physical height for recovering the distance is replaced with the groundtruth physical height. This supports our method can maintain the self-consistency during inference. From `K', we can see that, the accuracy increases if the predicted physical size for optimizing the distance is replaced with the groundtruth physical size. This shows there is no correlation, i.e., self-consistency, between the predicted projected corners and physical size.

\subsection{Comparisons on the KITTI benchmark}
We comprehensively benchmark MonoRCNN on the KITTI official test dataset~\cite{DBLP:conf/cvpr/GeigerLU12} in Tab.~\ref{tab:Test}, and show the advantages of our method compared with existing methods. From Tab.~\ref{tab:Test}, we can see: 

1) MonoRCNN achieves the SOTA accuracy. Existing image-only methods are not able to emulate the methods using extra depth input, while our method pushes the forefront of image-only methods~\cite{DBLP:conf/eccv/SimonelliBPRK20}, by $3.17/1.75$ in AP$_\text{3D}$ and $2.72/1.08$ in AP$_\text{BEV}$ on the easy and moderate subset, 
and surpasses those depth-based methods by $1.59/0.93$ in AP$_\text{3D}$ and $0.45/0.79$ in AP$_\text{BEV}$ on the easy and moderate subset. With only single frame as input, our method is even comparable with a video-based method~\cite{DBLP:conf/eccv/BrazilPLS20}. Notice that MonoRCNN uses a ResNet-50 backbone~\cite{DBLP:conf/cvpr/HeZRS16}, while ~\cite{DBLP:conf/iccv/Brazil019} uses a DenseNet-121 backbone~\cite{DBLP:conf/cvpr/HuangLMW17} and ~\cite{DBLP:conf/cvpr/ChenTSL20,DBLP:conf/eccv/LiZLC20} use DLA-34 backbones~\cite{DBLP:conf/cvpr/YuWSD18}. Though DenseNet-121~\cite{DBLP:conf/cvpr/HuangLMW17} and DLA-34~\cite{DBLP:conf/cvpr/YuWSD18} are more advanced than our ResNet-50~\cite{DBLP:conf/cvpr/HeZRS16}, MonoRCNN still outperforms~\cite{DBLP:conf/iccv/Brazil019,DBLP:conf/cvpr/ChenTSL20,DBLP:conf/eccv/LiZLC20}. We also emphasize that MonoRCNN significantly outperforms RTM3D~\cite{DBLP:conf/eccv/LiZLC20}, a keypoint-based method, though \cite{DBLP:conf/eccv/LiZLC20} uses additional images from right cameras for training. This supports that our distance decomposition is much better than the keypoint-based distance decomposition.

2) MonoRCNN is simple and efficient. MonoRCNN is an image-only method, thus is more simple and efficient than those depth-based and video-based methods. The monocular depth estimator in ~\cite{DBLP:conf/cvpr/DingHYWSLL20a,DBLP:conf/iccv/MaWLZOF19,DBLP:conf/eccv/YeDSLTFDW20,DBLP:conf/eccv/MaLXZZO20} uses a heavy ResNet-101 backbone~\cite{DBLP:conf/cvpr/HeZRS16}, and MonoRCNN runs \num{3} times faster than~\cite{DBLP:conf/cvpr/DingHYWSLL20a}, and \num{5} times faster than~\cite{DBLP:conf/iccv/MaWLZOF19,DBLP:conf/eccv/YeDSLTFDW20,DBLP:conf/eccv/MaLXZZO20}.

\begin{figure*}
\begin{minipage}{0.5\linewidth}
\includegraphics[width=1.0\textwidth]{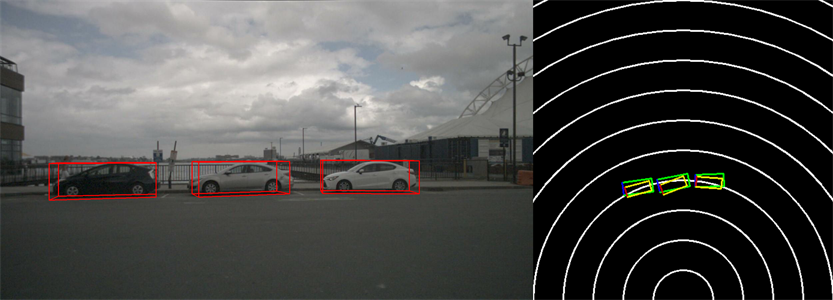}
\end{minipage}
\begin{minipage}{0.5\linewidth} 
\includegraphics[width=1.0\textwidth]{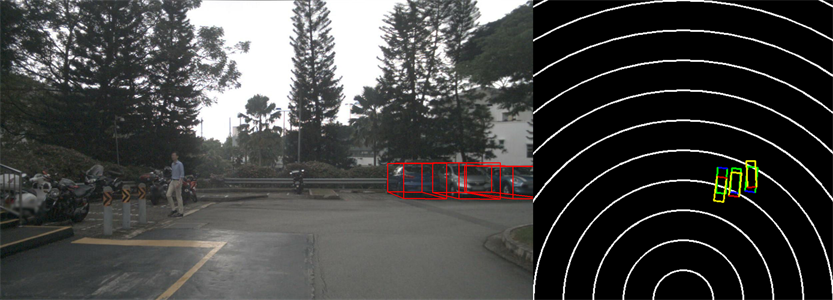}
\end{minipage}
\caption{\textbf{nuScenes Cross-Test Examples}. We visualize qualitative examples of MonoRCNN on the nuScenes~\cite{DBLP:conf/cvpr/CaesarBLVLXKPBB20} cross-test set. The 2D projections and bird's eye view results are shown as in Fig.~\ref{fig:Example_K}. Our model is only trained with the training subset of the KITTI val split~\cite{DBLP:conf/nips/ChenKZBMFU15}, and can generalize to images in the nuScenes~\cite{DBLP:conf/cvpr/CaesarBLVLXKPBB20} cross-test set with different camera intrinsics.}
\label{fig:Example_N}
\end{figure*}

\subsection{Cross-Dataset Test}
To evaluate the ability to generalize to images with different camera intrinsics, we conduct cross-dataset test by applying the model trained with the training subset of the KITTI val split~\cite{DBLP:conf/nips/ChenKZBMFU15} to the nuScenes~\cite{DBLP:conf/cvpr/CaesarBLVLXKPBB20} cross-test set. Since this paper is the first paper presenting cross-dataset test in monocular 3D object detection, we also provide the results of M3D-RPN~\cite{DBLP:conf/iccv/Brazil019} using its official model~\footnote{https://github.com/garrickbrazil/M3D-RPN} as a comparison. To focus on the accuracy of the distance prediction, we report the mean error of the distance prediction within different distance ranges for both methods in Tab.~\ref{tab:Cross}. The errors are calculated for recalled objects. The results show that our method achieves lower distance prediction errors. From errors in different distance intervals, we can see that the further the objects are, the better generalization of our method is. This is because our method reasons the distance by the local geometric variables of objects. We also visualize some qualitative examples in Fig.~\ref{fig:Example_N}, and we can see our method achieves accurate distance prediction. The pseudo-LiDAR-based methods~\cite{DBLP:conf/cvpr/WangCGHCW19,DBLP:conf/iccv/MaWLZOF19,DBLP:conf/eccv/YeDSLTFDW20} suffer from the difficulty of generalizing to images with different camera intrinsics, because it is difficult for monocular depth estimators to generalize to images with different camera intrinsics~\cite{DBLP:conf/cvpr/FacilUZMBC19}.

\begin{table}
\begin{center}
\begin{adjustbox}{max width=\textwidth}
\begin{tabular}{l | c | c c c}
\toprule[1pt]
\multirow{2}{*}{} &
\multicolumn{4}{c}{Distance prediction mean error (meters) \textdownarrow}  \\
 & $[0,  +\infty)$ & $[0, 20)$ & $[20, 40)$ & $[40, +\infty)$ \\
\midrule[0.5pt]
\cite{DBLP:conf/iccv/Brazil019} (T) & 1.26 & 0.56  & 1.33  & 2.73 \\
\cite{DBLP:conf/iccv/Brazil019} (N) & 2.75 & 1.04  & 3.29  & 10.73 \\
\midrule[0.5pt]
Ours (T) & 1.14 & 0.46  & 1.27  & 2.59 \\
Ours (N) & 2.39 & 0.94  & 2.84   & 8.65 \\
\bottomrule[1pt]
\end{tabular}
\end{adjustbox}
\end{center}
\caption{\textbf{Cross-Dataset Test} on the nuScenes~\cite{DBLP:conf/cvpr/CaesarBLVLXKPBB20} cross-test set. We show the mean error of the distance prediction within different distance ranges on different test datasets. `T' means the val subset of the KITTI val split~\cite{DBLP:conf/nips/ChenKZBMFU15}, and `N' means the nuScenes~\cite{DBLP:conf/cvpr/CaesarBLVLXKPBB20} cross-test set. All models are only trained with the training subset of the KITTI val split~\cite{DBLP:conf/nips/ChenKZBMFU15}.}
\label{tab:Cross}
\end{table}

%% file: Content/Materials.tex
\begin{figure*}
\begin{minipage}{0.5\linewidth}
\centering
\includegraphics[width=1.0\textwidth]{./Figure/Example_K_Val/000028}
\includegraphics[width=1.0\textwidth]{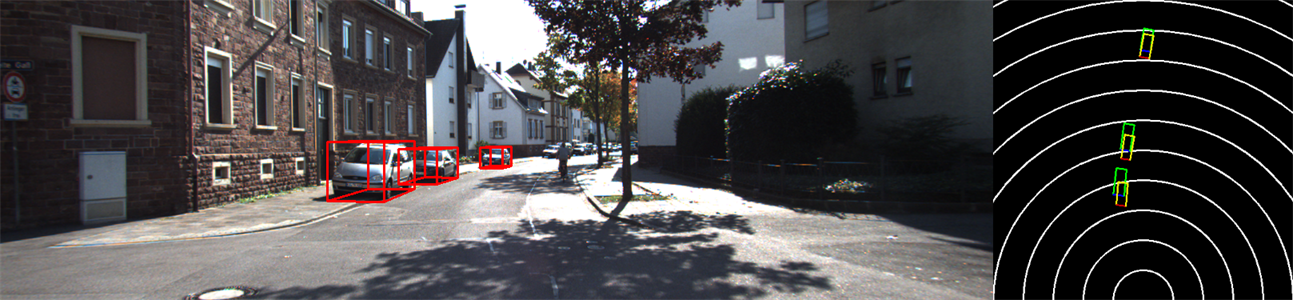}
\includegraphics[width=1.0\textwidth]{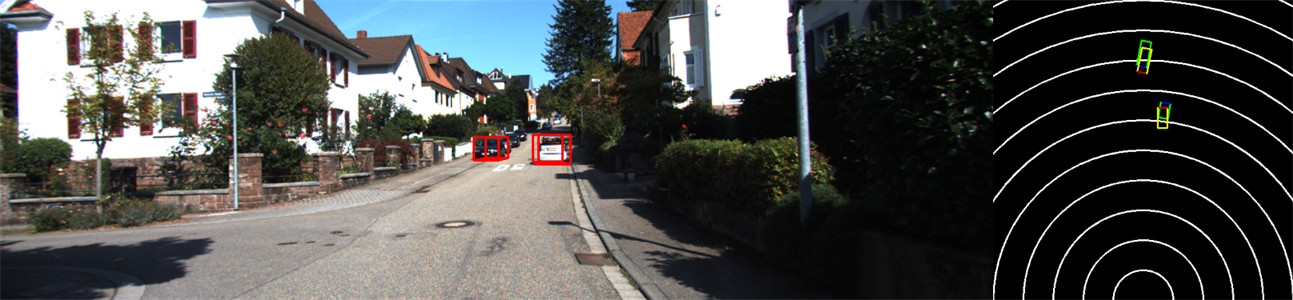}
\includegraphics[width=1.0\textwidth]{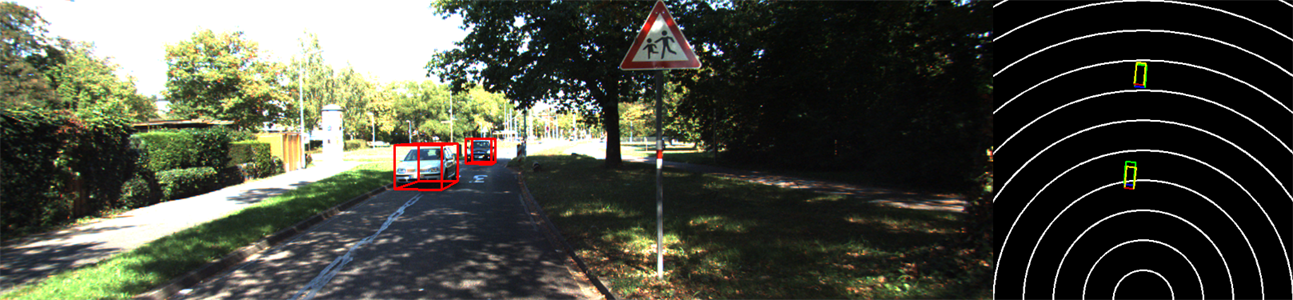}
\includegraphics[width=1.0\textwidth]{./Figure/Example_K_Val/000104}
\includegraphics[width=1.0\textwidth]{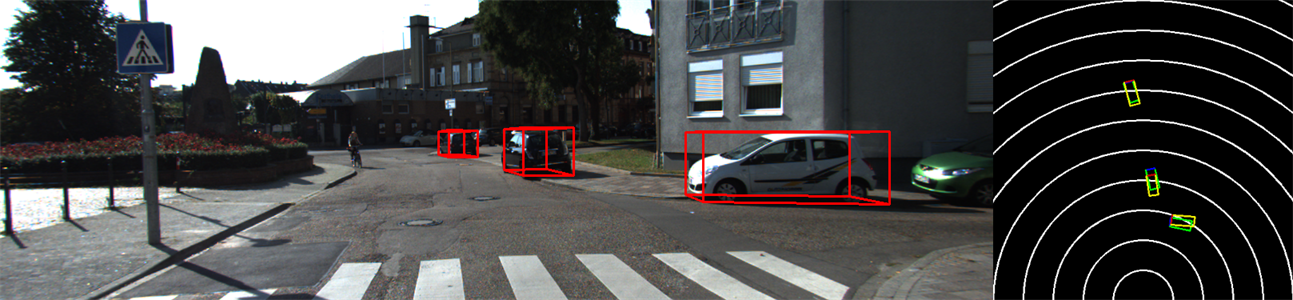}
\end{minipage}
\begin{minipage}{0.5\linewidth}
\centering
\includegraphics[width=1.0\textwidth]{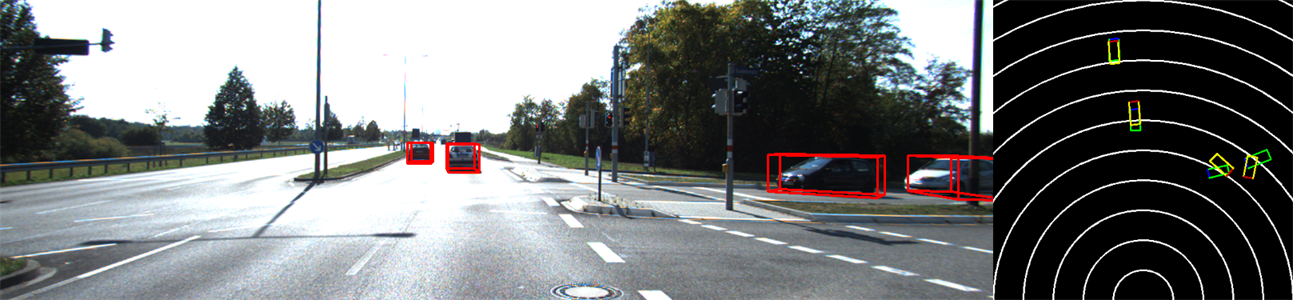}
\includegraphics[width=1.0\textwidth]{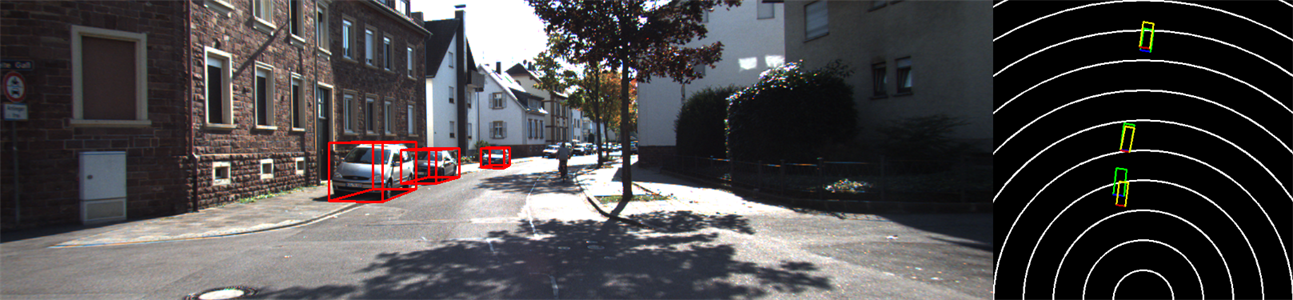}
\includegraphics[width=1.0\textwidth]{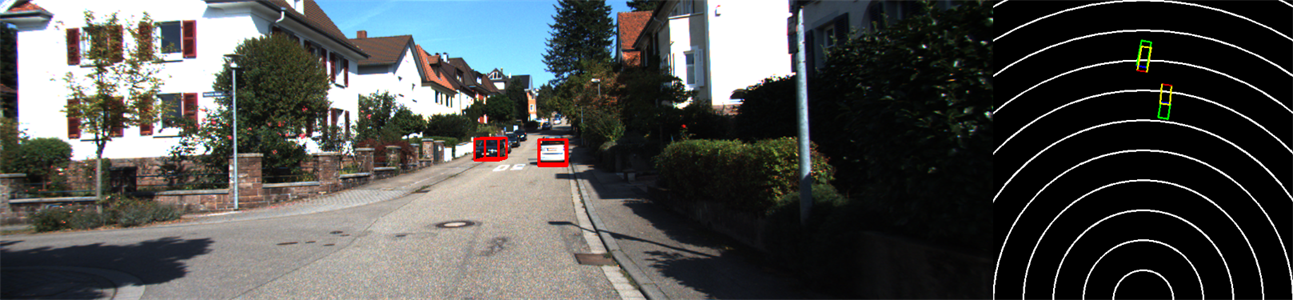}
\includegraphics[width=1.0\textwidth]{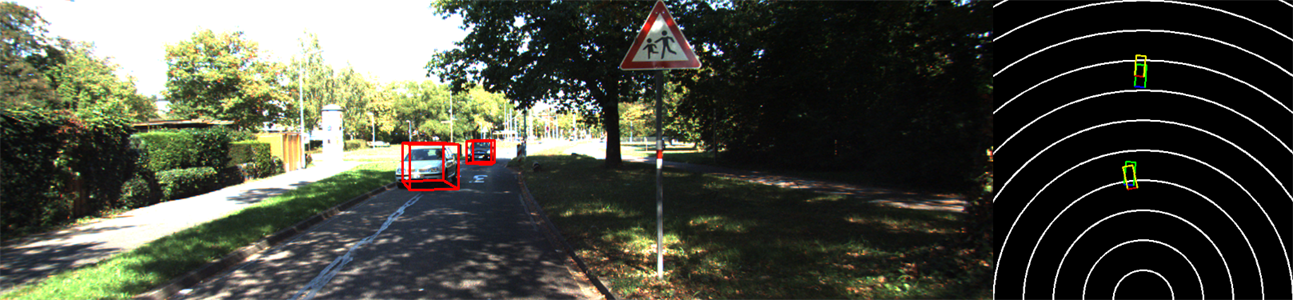}
\includegraphics[width=1.0\textwidth]{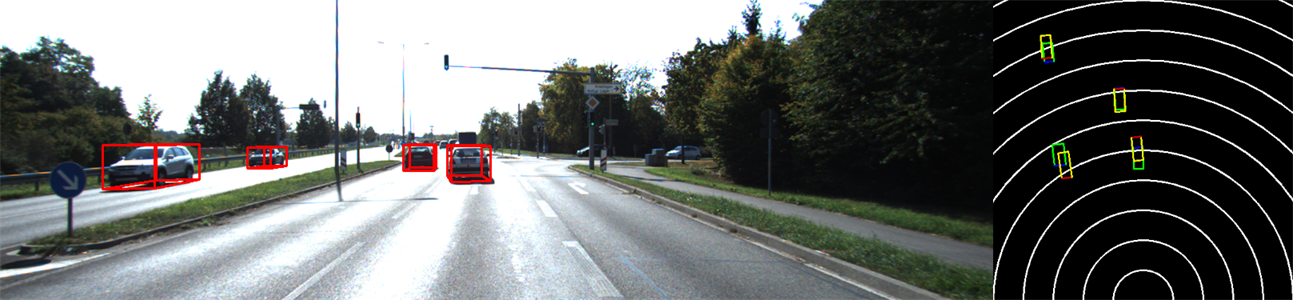}
\includegraphics[width=1.0\textwidth]{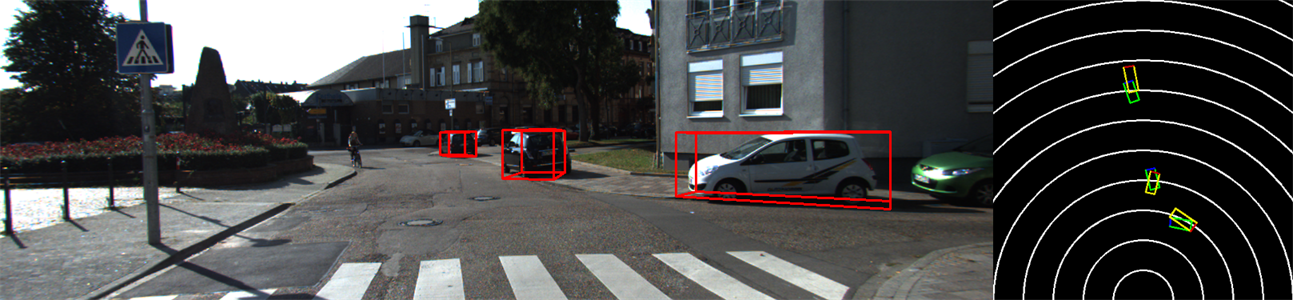}
\end{minipage}
\caption{\textbf{KITTI Val Examples}. We visualize qualitative examples of MonoRCNN (left) and M3D-RPN~\cite{DBLP:conf/iccv/Brazil019} (right) on the val subset of the KITTI val split ~\cite{DBLP:conf/nips/ChenKZBMFU15}. We can see our method is more accurate than M3D-RPN~\cite{DBLP:conf/iccv/Brazil019}. The \textcolor{red}{red} boxes in the image planes represent the 2D projections of the predicted 3D bounding boxes. The \textcolor{yellow}{yellow} / \textcolor{green}{green} boxes in the bird's eye view results represent the predictions and groundtruths of the 3D bounding boxes, respectively, and the \textcolor{red}{red} / \textcolor{blue}{blue} lines indicate the yaw angle of cars. The radius difference between two adjacent white circles is \num{5} meters. All illustrated images are not used for training.}
\label{fig:Example_K_MA}
\end{figure*}

\begin{figure*}
\begin{minipage}{0.5\linewidth}
\centering
\includegraphics[width=1.0\textwidth]{./Figure/Example_N/000117.png}
\includegraphics[width=1.0\textwidth]{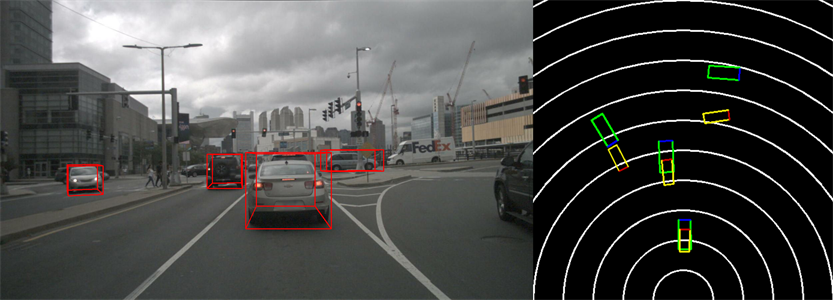}
\includegraphics[width=1.0\textwidth]{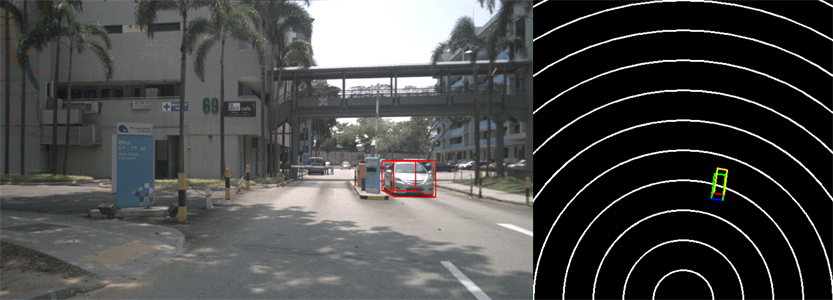}
\includegraphics[width=1.0\textwidth]{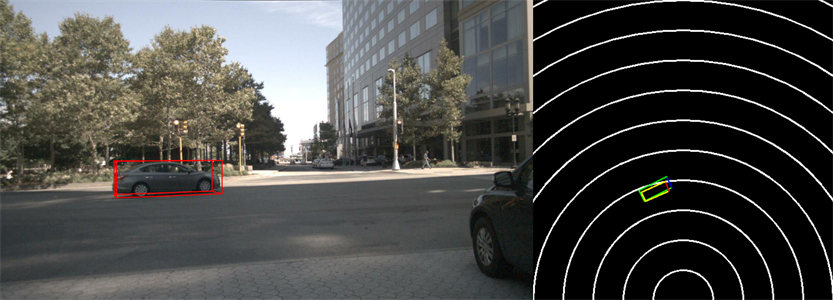}
\includegraphics[width=1.0\textwidth]{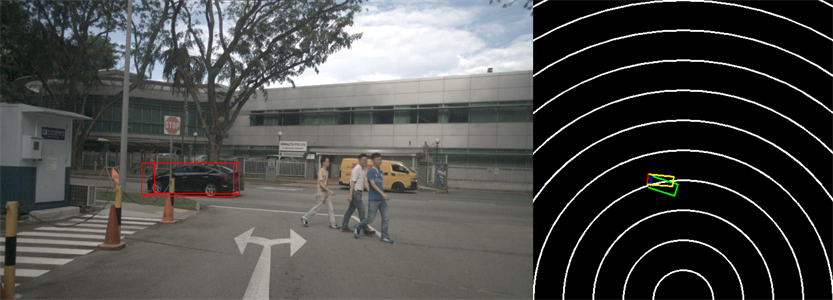}
\includegraphics[width=1.0\textwidth]{./Figure/Example_N/000108.png}
\end{minipage}
\begin{minipage}{0.5\linewidth} 
\centering
\includegraphics[width=1.0\textwidth]{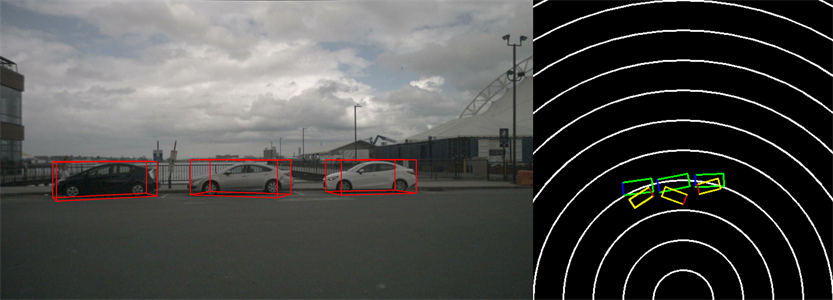}
\includegraphics[width=1.0\textwidth]{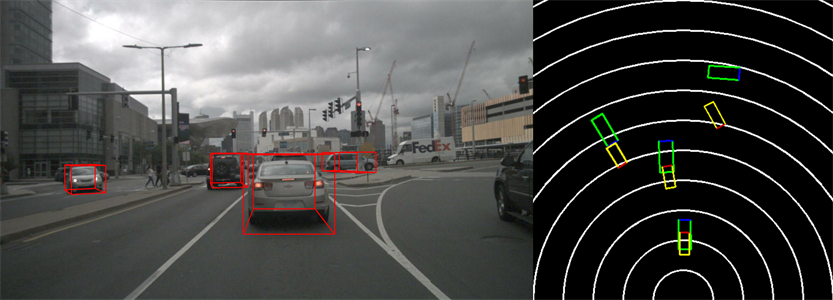}
\includegraphics[width=1.0\textwidth]{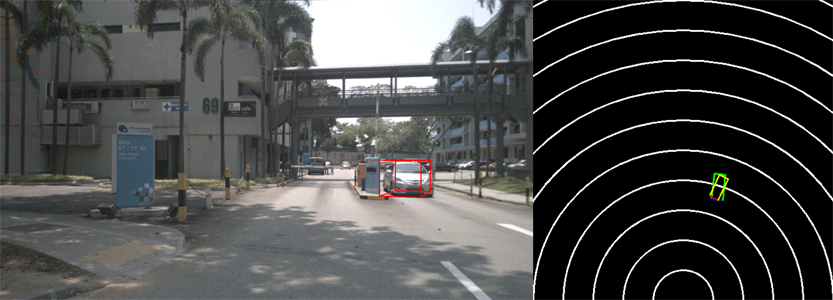}
\includegraphics[width=1.0\textwidth]{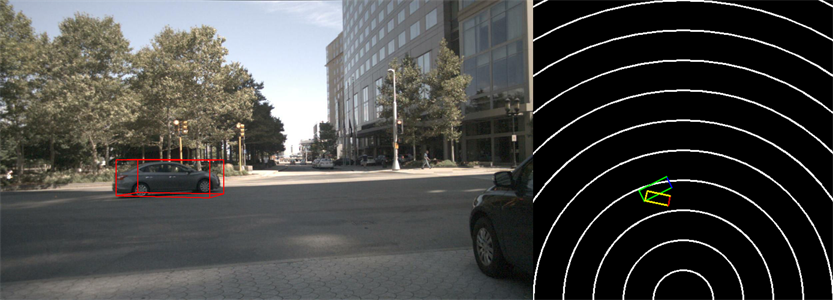}
\includegraphics[width=1.0\textwidth]{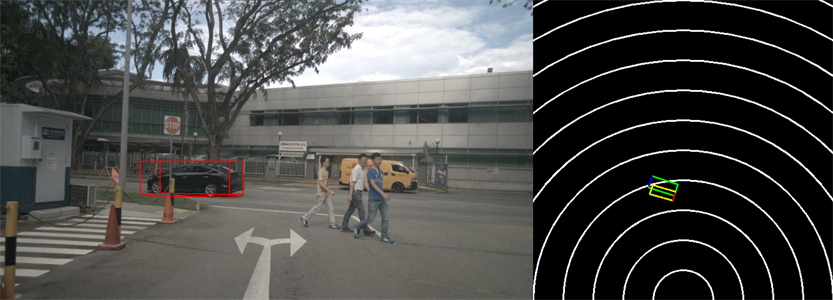}
\includegraphics[width=1.0\textwidth]{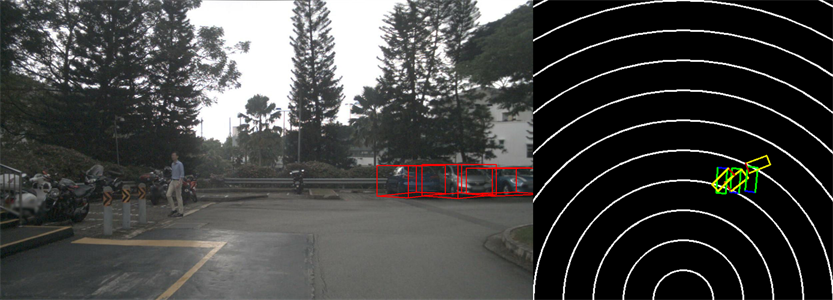}
\end{minipage}
\caption{\textbf{nuScenes Cross-Test Comparisons}. We visualize qualitative examples of MonoRCNN (left) and M3D-RPN~\cite{DBLP:conf/iccv/Brazil019} (right) on the nuScenes~\cite{DBLP:conf/cvpr/CaesarBLVLXKPBB20} cross-test set. We can see our method achieves more accurate distance prediction. The 2D projections and bird's eye view results are shown as in Fig.~\ref{fig:Example_K_MA}. All models are only trained with the training subset of the KITTI val split~\cite{DBLP:conf/nips/ChenKZBMFU15}.}
\label{fig:Example_N_MA}
\end{figure*}